# Automated architectural space layout planning using a physics-inspired generative design framework

Z. Li[1], S. Li[1], G. Hinchcliffe[2], N. Maitless[1], N. Birbilis[3]

[1]College of Engineering and Computer Science, Australian National University, Acton, A.C.T, 2601. Australia
[2]College of Arts and Social Sciences, Australian National University, Acton, A.C.T, 2601. Australia
[3]Faculty of Science, Engineering and Built Environment, Deakin University, Waurn Ponds, VIC. 3216. Australia

**Abstract**:

The determination of space layout is one of the primary activities in the schematic design stage of an architectural project. The initial layout planning defines the shape, dimension, and circulation pattern of internal spaces; which can also affect performance and cost of the construction. When carried out manually, space layout planning can be complicated, repetitive and time consuming. In this work, a generative design framework for the automatic generation of spatial architectural layout has been developed. The proposed approach integrates a novel physics-inspired parametric model for space layout planning and an evolutionary optimisation metaheuristic. Results revealed that such a generative design framework can generate a wide variety of design suggestions at the schematic design stage, applicable to complex design problems.

# 1. Introduction

In architectural design, the determination of a building's internal space layout is a crucial component of the schematic design phase. During this stage, the foundational spatial arrangement is conceptualised, setting the stage for subsequent spatial interactions and functional efficacy. Typically, architects initiate the space layout design by creating rough sketches or diagrams to delineate the positions and interrelationships of distinct functional areas, subsequently refining these into multiple design solutions. The meticulous planning of space layout, which outlines the internal spaces' form, size, and circulation patterns, directly influences the building's operational performance and economic outlay [1, 2]. Layout planning is recognised as a wicked problem due to its inherent complexity and variability [3]. This complexity tends to escalate, presenting a compounded challenge for human designers as the scale and intricacies of the project increase. Computational design and design automation techniques have been utilised extensively within the realm of architecture, offering significant time savings by streamlining repetitive tasks and thereby enhancing designer productivity [4-7]. This efficiency has paved the way for these technologies to be integrated more deeply into architectural practices. Consequently, it is a natural progression to employ these automated techniques to assist designers in the repetitive or complex task of space layout planning in architecture.

In recent years, generative design and automated generation of floorplans and space layout has garnered considerable interest, indicating a potential paradigm shift in design methodologies. [8-14]. Generative design is a design exploration method that utilises computational algorithms to automatically generate various design solutions that meet certain user-defined criteria (and limitations) [14-16]. With generative design, instead of directly interacting with the design product, the designer can set a design space and a series of design objectives for the computer algorithm to generate design solutions – helping designers explore larger design space within the computer aided design environment [14]. Popular generative design algorithms in architecture include cellular automata, shape grammar, evolutionary algorithms, and machine learning.

Cellular automata are effective generative design methods in architectural projects, particularly where there is a degree of repetition across numerous elements [17]. They have been applied in various sectors of the architecture industry, such as urban planning [18-20], and building methodology [21]. The generative design process using cellular automata is governed by a set of transition rules acting on a regular grid of cells. Herr and Kvan reviewed the application of cellular automata in architecture and proposed a responsive cellular automata-supported generative design process, which allows human control over a variety of generative design outcomes [17]. Moreover, the application potential of cellular automata in architecture has been demonstrated through the use of a generative model in the design of three-dimensional architectural forms for a high-density residential project [21]. However, it is important to acknowledge the limitations of cellular automata in architectural applications, as their rule-based, grid-dependent nature may not fully capture the continuous variability and complexity of real-world architectural forms and functions.

Some authors have applied shape grammar to generate floorplans of building. The generation process of shape grammar is basically the application of hard-coded rules of interaction upon a set of primitive shapes [22]. Once established, these rules can be systematically used to generate a multitude of designs, yielding an extensive array of results from straightforward initial rules. For instance, Mitchell and Stiny developed a parametric shape grammar that could generate floorplans of villa in the Palladian style [23]. In 1996, a generative shape grammar model for the design of English row-houses was developed [24]. Moreover, in 2009, Trescak et al. proposed a shape grammar framework which could automatically generate building floorplans and allow users to participate in the design process [22]. However, a limitation of this approach is that it may yield designs that are too complex or even impractical – however because it is based on fixed rules and primitive shapes, results are generally restricted to specific architectural styles.

Evolutionary algorithms are search algorithms that can be applied to find optimal or near-optimal solutions to complex problems by simulating nature's evolution process. Evolutionary algorithms are suitable for design optimisation and have been widely applied to explore creative designs in the schematic design stage [16]. As the most recognised form of evolutionary algorithms, genetic algorithms, initially proposed by Holland [25], are renowned for their robustness in navigating vast search spaces and have long been viewed as effective tools in search, design and optimisation [26, 27]. Owing to their stochastic nature, genetic algorithms facilitate the exploration of the design space and the generation of diverse design solutions with less dependence on extensive specialised domain knowledge. By performing multidirectional search in the search space (genes) and evaluating the individuals in the solution space (phenotypes), genetic algorithms may solve design and optimisation problems that have highly non-linear fitness functions; which are inherently difficult to solve through traditional optimisation methods. This makes genetic algorithms particularly suitable for the design and optimisation in engineering applications [28]. The application of genetic algorithms in architectural design and floorplan generation has been extensively explored by the research community over the last two decades [11, 29-35]. In 2006, Caldas developed an evolution based generative design framework named GENE_ARCH aimed at fostering sustainable architecture [29]. In 2013, Rodrigues et al. proposed an approach based on a hybrid evolutionary strategy for addressing multi-level space allocation problems [31]. Song et al. (2016) proposed an alternative evolutionary-based architectural design method in the design of apartment buildings [32]; whilst more recently, Zhang et al. (2021) developed a parametric generative algorithm to automatically generate energy efficient design of residential buildings [33]. The effectiveness of genetic algorithms significantly hinges on the sophistication of the parametric model and the well-definition of objective functions. Consequently, while the studies cited have successfully demonstrated the generation of valid and usable outcomes, it's important to recognise that certain approaches may yield results that, while practical, could be somewhat over-simplified or exhibit limited flexibility in terms of design variability.

Recent advancements in generative machine learning, particularly variational autoencoders (VAEs) [36] and generative adversarial networks (GANs) [37-39] have been transformative in various engineering design fields, including topology design and optimisation [40-42], materials design [43-45] , and architecture [46-50]. Typically, in most studies, these generative models are trained to create designs using image-based data for architectural space planning. For instance, Huang and Zheng introduced a pix2pix GAN model in 2018 that could learn from and generate layout images of architectural plans [48]. Similarly, in 2019, Chaillou developed a generative model called ArchiGAN that could be applied in apartment building designs [50]. Nauata et al. applied relational GANs to transform constrained graphs to realistic house floorplan layouts [46]. More recently, Jia et al. (2023) present a dual-module approach that leverages GANs for multi-style interior floor plan design, demonstrating enhanced versatility and precision in the generated layouts [51]. A major advantage of these methods is their ability to produce a large volume of solutions in a short time once the model is adequately trained. However, their application in architectural and space layout design still faces several challenges. One primary issue is data sparsity due to the difficulty in acquiring standardised, well-annotated, and high-quality data, which is both challenging and time-consuming. Another concern is the feasibility and creativity of the generated outcomes, models may yield undesirable results that simply mimic training data [52]. Moreover, many studies using image-based outputs often lack the parametric details necessary for direct professional application, limiting their practical utility.

This study introduces a generative design framework that combines a novel physics-inspired parametric model with an evolutionary optimisation technique to generate internal space layouts of buildings. For a set of pre-defined design constraints and objectives, the proposed generative design framework is capable of autonomously generating and optimising a variety of design solutions. The innovation of the parametric model lies in its use of 'fields', a concept borrowed from physics, to address layout planning. The rooms within the plan are conceived as sources of virtual 'fields', with their magnitudes diminishing with distance from their 'centres of mass'. The space allocation to rooms is thus influenced by the relative strength of these fields, which transforms space layout planning into a competitive space allocation problem, whereby the rooms in a floorplan compete with each other to occupy area inside the building envelope by adjusting their positions and 'field' parameters. A pathway generation algorithm was developed to facilitate the generation of circulation patterns in the space allocation results. To acquire high quality design solutions, finding optimal combinations of parameters are sought, in order to serve the parametric model. Herein a multi-objective genetic algorithm called the non-dominated sorting genetic algorithm (NSGA-II) [53] was applied to search and explore optimal solutions. Following the presentation of the generative design framework, two case studies where the proposed generative design framework was applied, are presented - showing the efficacy and applicability of the proposed generative design framework in architectural projects of various scales.

## 2. Generative design framework

In this study, a generative design framework that can automatically generate the internal layout of buildings is presented. For a series of predefined design constraints and design objectives, the proposed model could generate a variety of design solutions through an evolutionary search process. The aim of the proposed generative design method is to utilise computational resources in order to assist human designers by generating a variety of design solutions that may be further improved in the schematic design stage. The generative design framework proposed consists of two components:

1) A physics-inspired parametric model that can be applied to generate diverse design solutions of internal space layouts of buildings, and,

2) An evolutionary algorithm that can search and explore a variety of optimal design solutions for the parametric model (herein NSGA-II is applied).

A schematic of the generative design framework employed herein is depicted in Figure 1.

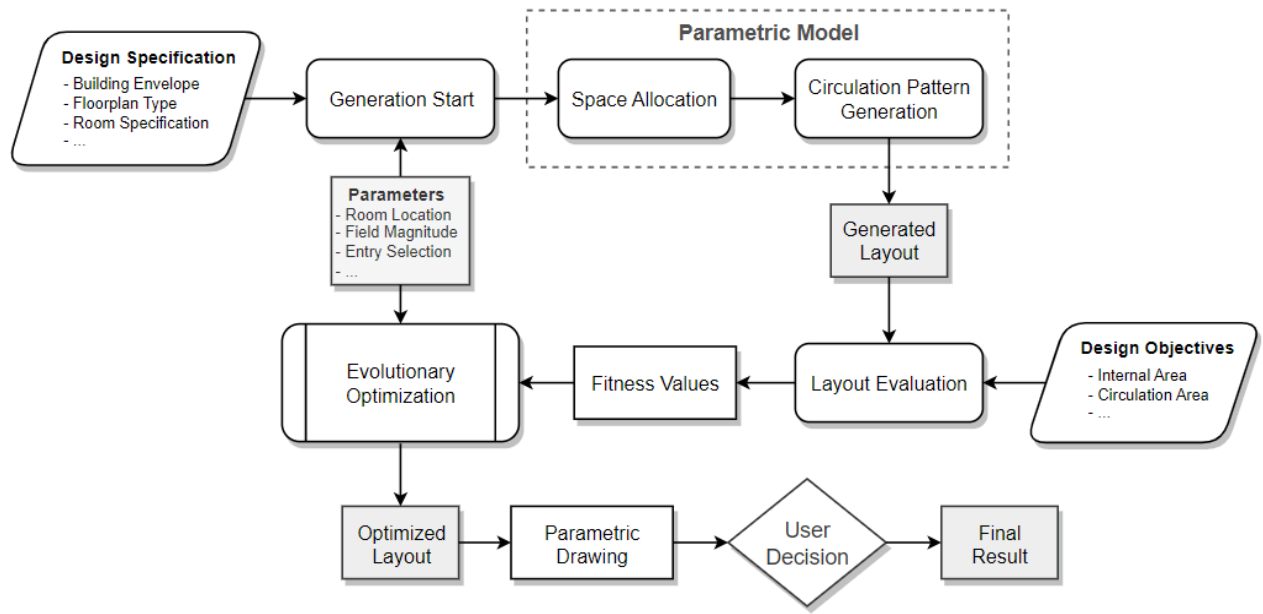

**Figure 1**. Workflow of the generative design framework employed in the study herein.

Prior to any layout generation, the design area inside a building envelope is decomposed into numerous grid cells. The distribution of these cells among various rooms leads to the creation of distinct space layout plans. This physics-inspired parametric model generates a space layout in two steps. For a given set of design specifications and parameters, the model first runs an algorithm to carry out cell allocation. With each room generating a virtual 'field', this process involves calculating field magnitudes at locations of all the grid cells for the rooms and allocating cells to the rooms according to their field magnitudes. Then, a circulation pattern generation algorithm developed based on Dijkstra's algorithm [54] is applied to draw a circulation pattern on the space allocation result; which indicates the pathways from the entry of the whole building to all the rooms. The generated space layout is then evaluated according to a set of predefined design objectives such as internal area, circulation area etc. All the generated space layouts (solution candidates) are iteratively optimised by an evolutionary algorithm. Herein the optimisation is implemented with an evolutionary solver called Wallacei, which is an multi-objective analytic and optimisation engine for Grasshopper 3D [55]. The generative design framework built in Grasshopper 3D is shown in Figure 2. The final outputs of the proposed generative design framework are parametric designs that can be directly utilised with common design tools / software. The components within the generative design framework are elaborated further below, whereby the framework has been developed as a system to harness user input towards generated results.

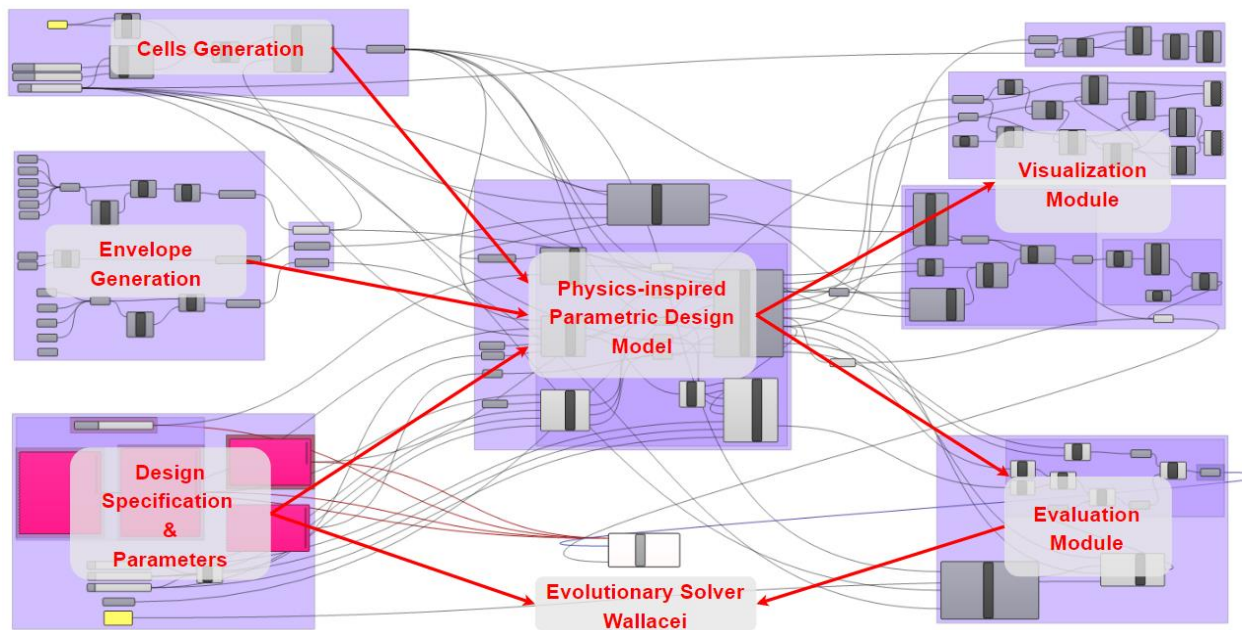

**Figure 2.** Generative design framework developed in Grasshopper 3D

### 2.1 Parametric model

A space allocation problem (SAP) in architecture considers the determination of the locations and dimensions of rooms in a two-dimensional space under certain design constraints and objectives [56]. Herein the space allocation problem was considered as a cell allocation problem. As noted, prior to layout generation, the design area inside a building envelope is decomposed into a finite number of equal-sized grid cells. The proposed parametric model employs a method that applies virtual fields to determine the allocation of cells. In this method, each room in the floorplan has a 'mass' and could generate a virtual scalar field in the xy-plane, whose magnitude is inversely correlated to the distance from the 'mass centre'. The allocation of cells to the rooms is determined by following rules:

(1) A room's field on a cell is termed 'active' if the field magnitude surpasses a predefined threshold parameter $\delta$;

(2) A room is permitted to claim a grid cell if the cell falls under the room's active field—this, in conjunction with (1), guarantees that the generated rooms can only occupy a limited number of areas;

(3) When a cell falls under the active fields of multiple rooms, the cell's space is assigned to the room exerting the strongest field.

As the field magnitudes are calculated mathematically, by applying different mathematical formulas, virtual fields of various geometrical forms can be generated. By adjusting the parameters of the mathematical formulas, one may thus control the position and geometrical forms of the generated rooms.

While the space allocation result indicates the position and shape of each room, to generate feasible space layouts for buildings, a circulation pattern generation algorithm was developed to facilitate this process. The circulation pattern generation algorithm was developed based on Dijkstra's shortest path algorithm, which is a popular algorithm for finding shortest paths between nodes in a graph [54]. A weighted graph is therefore constructed for each space allocation result where the cells are represented as nodes. These nodes are linked to their adjacent nodes by edges of varying lengths. The generated circulation pattern connects the entrance of the floorplan and the entrances of all the generated rooms. Although this pattern is derived from the room allocation results, the iterative nature of the entire generation process means that the circulation pattern also influences the final positioning and form of the rooms. Figure 3 presents a comparative illustration featuring the input design specifications alongside the optimised output generated by the generative design model developed in this study.

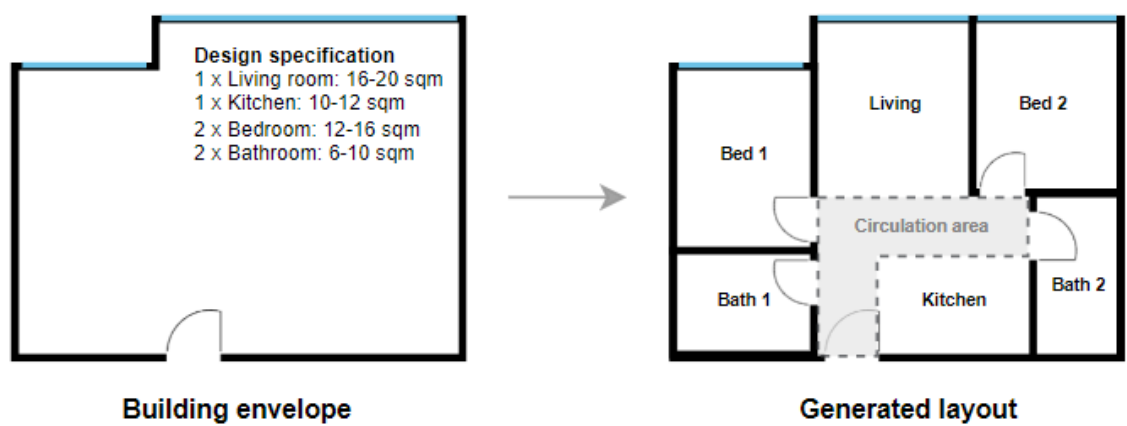

**Figure 3.** Illustration of the design specification and generated space layout of the generative design model developed in this work.

### 2.1.1 Design specification and parameters

The design specifications encompass the building or apartment's envelope, entrance, fenestration, along with the number and target dimensions for each room type. These parameters are established at the outset of the generative design process to delineate the search space and design constraints for the layout generation. Notably, room dimensions can be specified as ranges to allow for flexibility; for instance, a bedroom might be designated to have a fixed length of 4 meters but a variable width of 3 to 4 meters, resulting in an area that could range from 12 to 16 square meters. Within the parametric model, factors such as room positions and their field strengths are subject to optimisation. In the context of evolutionary optimisation, these factors act as the 'genes' of a candidate solution, shaping the geometric and functional attributes of the final design.

### *2.1.2 Computational complexity*

The computational complexity of the parametric model for space allocation is primarily influenced by the two phases: cell allocation and circulation pattern generation. During the cell allocation phase, each cell undergoes a computation of field magnitudes relative to each room, followed by identifying the maximum value among these magnitudes to determine the cell's allocation. This results in a computational complexity of $O(MN)$ for each potential design solution, where $M$ is the number of rooms, and $N$ is the number of grid cells. Following this, the circulation pattern generation phase employs an algorithm developed based on Dijkstra's shortest path algorithm, which has a computational complexity of $O(N^2)$. Consequently, the overall computational complexity of generating the space layout is determined by the sum of the complexities from both the cell allocation and circulation pattern generation phases, $O(MN) + O(N^2)$. However, since the value of $N$ is significantly larger than $M$, the complexity is ultimately dominated by $O(N^2)$.

### *2.1.3 Grid cell generation*

Grid cell generation involves the decomposition of the interior area of a building envelope into a finite number of uniformly sized grid cells, which serve as the smallest units in space allocation. The size of the cells affects the search space and the 'resolution' of the generated space layout. By decomposing the design area into smaller grid cells, the model can generate space allocation results with higher resolution. This allows the model to fine-tune the position and geometric form of the generated rooms to better accommodate the building envelope and design specification, thereby improving the performance and diversity of the generated space layouts. However, since the size of an envelope is fixed, indiscriminately smaller grid cells result in a process of space allocation and so-called circulation pattern generation also increasing significantly. As the space layout generation process has a computational complexity of $O(N^2)$, the algorithm's time complexity will rise quadratically as grid cell dimensions diminish. Furthermore, smaller grid cells also expand the search space, potentially affecting the optimisation algorithm's efficiency in discovering optimal solutions. Hence, selecting a grid cell size that balances detail and computational efficiency is vital, and should be tailored to the demands of the specific architectural project in question.

### *2.1.4 Space (cell) allocation*

The shape and dimension of a generated room in space layout are affected by the shape and magnitude of its fields. Employing diverse mathematical functions and parameters allows for the creation of spatial layouts featuring a range of geometric configurations. Figure 4 illustrates this concept by displaying three distinct field types, each shaped by different mathematical functions, with the depth of colour representing the field magnitudes across the grid cells.

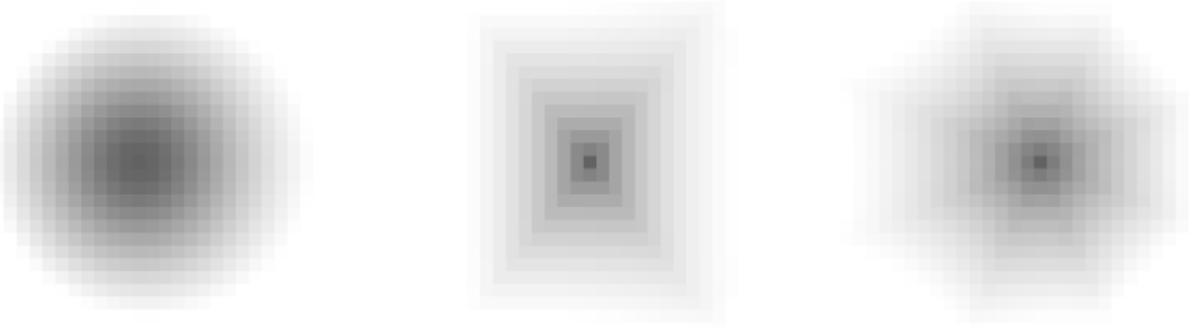

**Figure 4.** Three fields generated on the same design space (rendered in Rhino 7). The shapes of fields are respectively: circular, rectangular, and octagonal.

Given that rectangular rooms are typically more prevalent and practical in floorplan designs, the space allocation mechanism presented here is demonstrated using rectangle-shaped 'fields.' Originating from the 'mass centre' $(x_0, y_0)$ of the room, the field extends rectangularly in both the X and Y directions. The magnitude of this rectangular field $f$ at any given point $(x_1, y_1)$ is calculated as:

$$f = \frac{m_x * m_y}{|x_1' - x_0| + |y_1' - y_0| + \varepsilon} \quad \text{(Eqn. 1)}$$

$$x_1' = m_y * (x_1 - x_0) * \cos(t) - m_x * (y_1 - y_0) * \sin(t) + x_0 \quad \text{(Eqn. 1.1)}$$

$$y_1' = m_y * (x_1 - x_0) * \sin(t) + m_x * (y_1 - y_0) * \cos(t) + y_0 \quad \text{(Eqn. 1.2)}$$

where the variables $m_x$ and $m_y$ are the mass parameters that affect the magnitude of the field in the X and Y directions, and the variable $t$ is a parameter that controls the rotational angle of the field. When $t = \pi/4$, the field assumes an upright rectangular orientation. The variable $\varepsilon$ is a (negligible) positive value introduced to avert any division by zero errors in the calculations.

The magnitude of a room's field on a grid cell is calculated according to their relative locations and its mass parameters $m_x$ and $m_y$. When a cell is under the fields of multiple rooms, the cell is allocated to the room that exerts the largest field upon it - noting that the area in a cell will only be allocated when the magnitude of the field at its location is larger than a preset threshold value $\delta$. Therefore, while the field's theoretical reach may extend indefinitely, in practice, each room is constrained to occupy only a finite number of cells. By fine-tuning the mass parameters $m_x$, $m_y$, and $\delta$, it is possible to control the maximum extent of cells a room can span both vertically and horizontally.

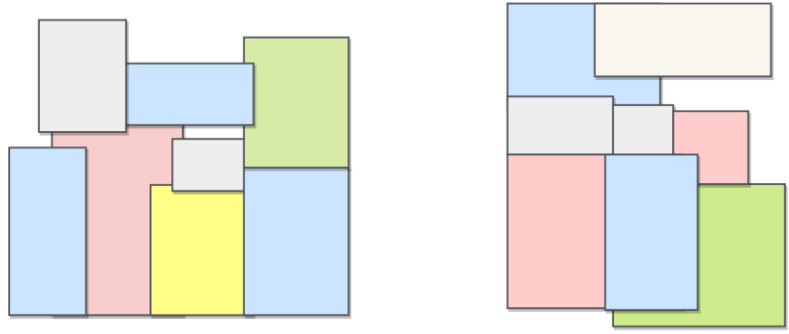

**Figure 5.** Examples of irregular space layouts generated using the bounding box method.

The field-based space allocation mechanism offers several key advantages. One of the principal advantages is that it allows the existence of conflicts and overlaps between rooms during optimisation and provides a concise way to resolve them. In many prior studies, the rooms within floorplans of generative design projects are commonly depicted as bounding boxes. The optimisation of these designs is typically conducted by directly manipulating the bounding boxes' position, length, and width, which can inevitably result in conflict areas and overlaps within the floorplans. There are typically two methods to address the issue of overlapping areas: (a) Assigning different priority levels to each room, and the overlapping areas are allocated to the rooms with higher priorities; (b) Implementing an additional algorithm to reallocate the overlapping spaces among the rooms. The first approach tends to produce space layouts with stacked bounding box configurations, as depicted in Figure 5. Whilst the second method that employs an algorithm to redistribute overlapping areas is usually computational costly and can generate space layouts with irregularly shaped rooms.

In our method, rather than altering room dimensions directly, the optimisation algorithm adjusts the generated rooms' positions and dimensions by modifying their field parameters. To prevent overlapping and irregular outcomes, a straightforward solution is to set the overlapping area as one of the optimisation objectives. Since each room is given a range of shape variability, as rooms continuously adjust their shapes and positions to compete for cell occupancy, overlapping areas can be effectively avoided and even eliminated during the iterative optimisation process.

Another notable advantage of our model is its ability to generate rooms with diverse dimensions. Room dimensions are not rigid but are instead defined within ranges, allowing the mass parameters $m_x$ and $m_y$ to be optimised within these predefined limits. Consequently, the dimensions of the generated rooms can vary within a certain range. For instance, the model can specify that a living room's width should be between 3 and 5 meters and its length between 4 and 6 meters. Depending on the optimisation process, the area of a living room could vary between 12 to 30 square meters. This variability in room dimensions grants the parametric model the flexibility to produce space layouts in a wide array of geometric shapes, thereby greatly enhancing the versatility of the generative design framework.

## 2.1.5 Circulation pattern generation

The circulation system is often considered as the skeleton of the floorplan which helps communicate and organise the functional elements of the building [57]. In our study, we developed a circulation pattern generation algorithm based on Dijkstra's shortest path algorithm [54], tailored to overlay circulation patterns onto the results of space allocation. The Dijkstra's shortest path algorithm is one of the most popular algorithms for finding the shortest path between nodes in a weighted graph with non-negative edges, which could find the shortest paths between a source node and every other node in the graph [54]. It has been widely applied to solve network routing problems in various fields such as digital map, robotic path etc.

Our proposed method for generating pathways involves a two-step process. The first step involves selecting an entrance for each room in the space allocation results. The second step involves determining the shortest paths from the floorplan's main entrance (the source node) to the entrances of all the generated rooms. Figure 6 showcases a simplified example of this methodology.

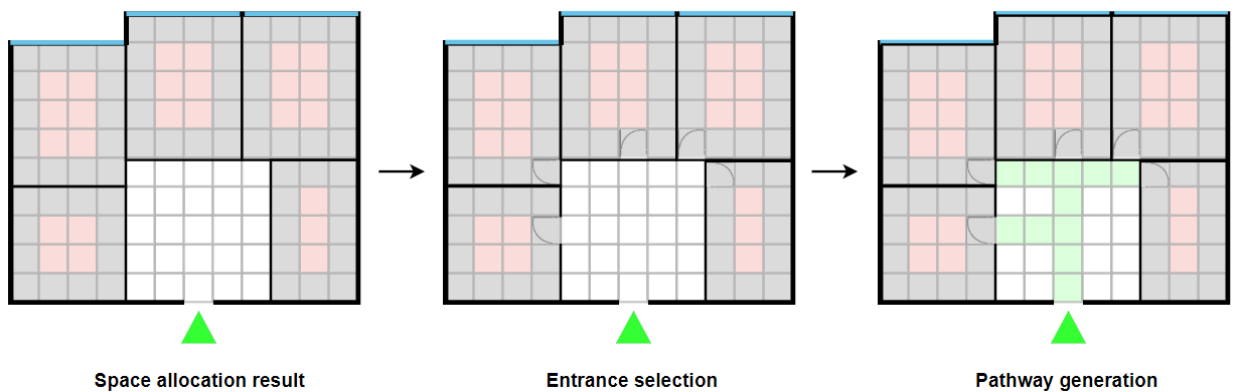

**Figure 6**. Illustration of the process of circulation pattern generation.

Leveraging Dijkstra's algorithm, we have incorporated a path shortening feature to enhance the connectivity and efficiency of the resultant circulation pattern. The first step in path generation is to define the weighted graph for the space allocation result. This graph is constituted of nodes interconnected by edges of various lengths [58]. For our purposes, the nodes represent a selected group of cells designated for pathway creation. We permit the formation of pathways on all unallocated cells (those not assigned to any room) and on the boundary cells (the peripheral cells of each room, as demarcated by the shaded cells in Figure 6 and Figure 7).

In this model, the weighted graph is designed to exclude the interior cells of the rooms, ensuring that the circulation pattern circumvents rather than traverses these spaces. From the boundary cells of each room, one cell is selected as the entrance point, whose selection is also a parameter for optimisation. As depicted in Figure 7, the interconnection between the adjacent nodes constitutes all the edges in the graph. The edge length varies depending on the type of the connected cells (nodes). The length assigned to each edge depends on the type of cells it connects: the edge between two free cells is set at a length of 2, between a free cell and a boundary cell at 5, and between two boundary cells at 10. This gradation ensures that the algorithm prioritizes free cells when plotting circulation pathways, thereby steering clear of the areas designated for rooms.

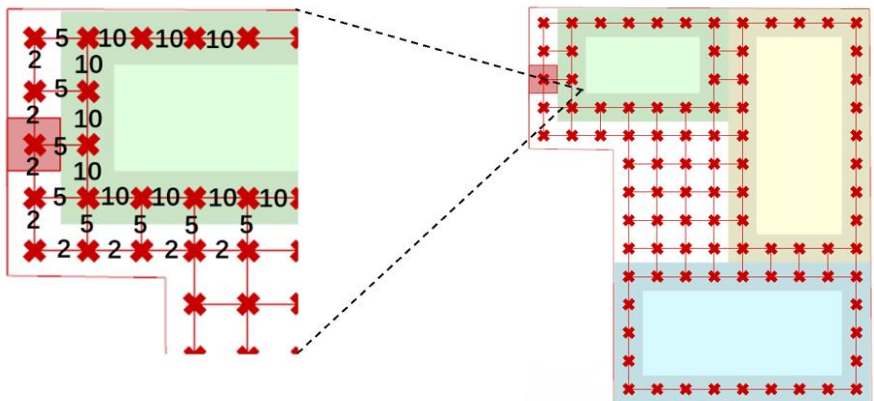

**Figure 7**. The weighted graph generated on the space allocation result, where the numbers in the enlarged graph indicate the lengths of edges.

The pseudocode for the PathGeneration algorithm is as follows:

```
1: function PathGeneration (Graph, source):
2:     for each node v in Graph:           //initialisation
3:         dist[v] = infinity              //the initial distance of vertex v is set to infinity
4:         p_node[v] = none                //the previous node of v in the shortest path
5:         p_edge[v] = none                //the length of the previous edge of v in the shortest path
6:     dist[s] = 0                         //the distance of source node is set to 0
7:     Q = the set of all nodes in Graph
8:     G = the set the all the entrance nodes in Graph
9:     while Q is not empty:               //main loop of the algorithm
10:        u = node in Q with shortest distance
11:        remove u from Q
12:        if u in G:      //when u is entrance node, shorten all the previous edges in its shortest path
13:            for each node w in the shortest path of u:
14:                p_edge[w] = 0.8 * p_edge[w]
```

The algorithm explores all the nodes in the graph one by one and calculate their distances to the source node iteratively. While the path generation algorithm is based on Dijkstra's algorithm, it is differentiated by a path shortening feature. As outlined in steps 12-14 of the PathGeneration algorithm, upon identifying an entrance node, the edge lengths along the shortest path from this node to the source are reduced by 20%. Consequently, as the algorithm seeks paths from the source node to all entrance nodes, it will, by preference, retrace the same paths as those identified for previously explored entrance nodes. This strategy aims to minimise the total area dedicated to circulation patterns and enhance the overall connectivity of the layout. Figure 8 illustrates a comparison between two circulation patterns: one generated using the standard Dijkstra's shortest path algorithm (on the left) and the other using the modified PathGeneration algorithm (on the right).

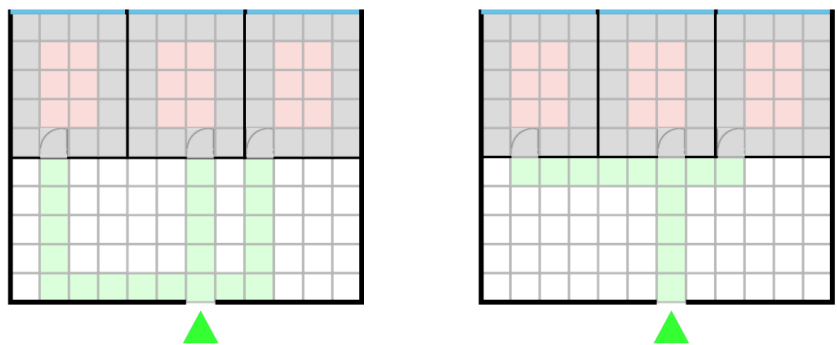

**Figure 8.** Illustration of the effect of the path shortening mechanism. The generation results of Dijkstra's shortest path algorithm (left) and the PathGeneration algorithm (right).

One issue with using Dijkstra's algorithm for circulation pattern generation is its handling of situations where multiple shortest paths are available to a newly discovered node. Dijkstra's algorithm typically connects to the first node in the queue, but since the queue order is random, the path selection is likewise random. This can lead to a layout with branching pathways to various entrances. However, the PathGeneration algorithm is designed to favour paths that share more nodes with previously established pathways, effectively reducing the randomness in path selection. Despite using the same weighted graph, the PathGeneration algorithm yields a more consolidated network of paths, as shown in Figure 8 (right). The paths from each entrance to the source node might be the same length as those produced by Dijkstra's algorithm (left graph), but the aggregate path length in the PathGeneration algorithm (right graph) is significantly shorter.

When generating floorplans within a fixed envelope, because the rooms will compete to occupy all the free cells, the outcomes from both algorithms may appear quite similar. However, in cases with non-fixed envelopes, such as house floorplans where the total length of pathways is considered as important design objective, the path shortening feature of the PathGeneration algorithm can create more efficient and cohesive circulation patterns. Figure 9 showcases circulation patterns for two single-story house plans generated using the proposed method, illustrating the effectiveness of the PathGeneration algorithm.

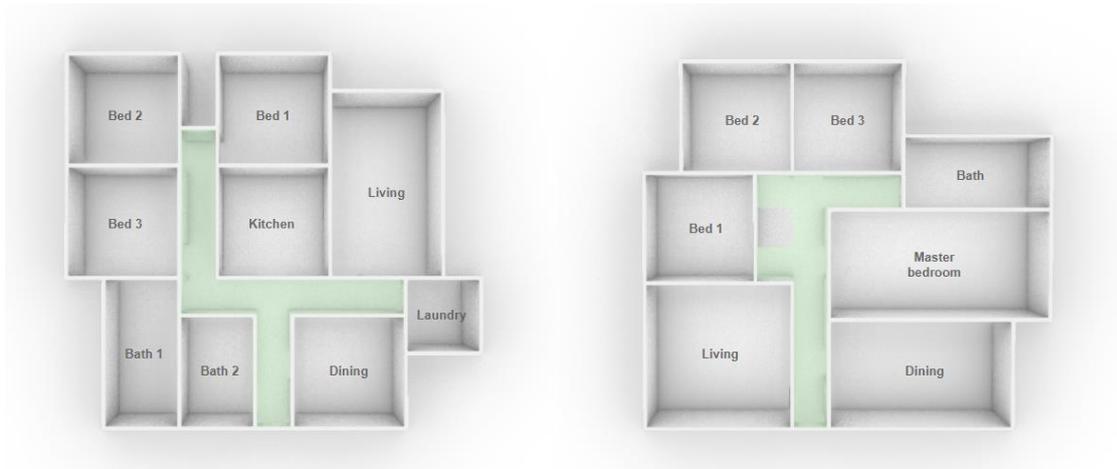

**Figure 9.** The space allocation and circulation pattern of two houses generated by the proposed generative design framework (and rendered in Rhino 7).

### 2.2 Evolutionary optimisation

The internal layout design of building is an inherently complex process involving a comprehensive analysis of various design objectives. Evaluating these objectives often involves subjective judgment and can deal with criteria that are not well-defined, particularly when it comes to aesthetic appeal and style, which are challenging to assess through automated means. In this context, we propose the use of multiple fitness values to quantitatively evaluate the generated spatial layouts, for aspects that include internal area, circulation pattern, and fenestration. The generation of optimal design solutions was through an evolutionary searching and optimisation process.

#### 2.2.1 Genetic algorithm

As the most recognised form of evolutionary algorithm, genetic algorithms are search algorithms that have long been viewed as effective tools in search, design and optimisation [26, 27]. These algorithms mimic the process of natural selection, sustaining and evolving a pool of candidate solutions, often referred to as the "population," to conduct a comprehensive, multidirectional search. Each candidate within this population is characterised by a genotype—its genetic makeup—and its phenotype—the manifested design. In the context of this study, the genotypic representation of an individual design solution comprises the parameters of the parametric model, while the phenotypic representation is the resultant space layout design. During the process of evolutionary optimisation, a series of fitness functions are applied to assess the viability of each candidate solution. Addressing the complexities of multi-objective optimisation, we utilized the non-dominated sorting genetic algorithm II (NSGA-II), a robust algorithm celebrated for its efficiency in identifying a spectrum of Pareto-optimal solutions within a single simulation run [59]. This study implemented the NSGA-II algorithm via the Wallacei evolutionary solver, an advanced multi-objective analytic and optimisation tool available within the Grasshopper 3D environment [55].

### 2.2.2 Design specification and parameters

As noted, the solution candidates were generated from a parametric model that was characterised by the design specification and associated parameters. Whilst the design specifications are predefined variables specifying the search space conditions, the design parameters are the 'genes' of the solution candidates, determining the geometric form and performance of the solution candidate. These design parameters are being iteratively optimised during the generative design process. The specifics of the design specifications and design parameters are detailed as follows.

Design specification:

1) An enclosed envelope curve that defines the boundary of the design space. In apartment space layout design, the fenestration of the building is also indicated on the envelope.
2) The entrance points of the envelope. Before the generation, one or multiple cells are defined as entrance candidates of which one cell is selected as the entry point of the whole floorplan. The selection of entrance is also a parameter for optimisation.
3) The number and target dimensions of each type of rooms. The target dimensions of rooms are defined in ranges, which also specify the optimisation range of the mass parameters of the rooms. For example, when the target width of a room is between 3 to 5 meters, the optimisation range of its mass parameter in the x direction is set as $m_x \in [3, 5]$. By defining the target dimensions in ranges, the search space for the optimisation algorithm is significantly increased, which improves the generative design framework's flexibility and creativity.
4) Any adjacency requirements between the rooms. For example, the kitchen and dining room are usually next to the living room in an open-plan style design.

Design parameters (genes):

1) The selection of entry point for the whole floorplan.
2) The coordinates of the mass centre of each room in the xy plane, $(x_0, y_0)$.
3) The mass parameters of the rooms. When applying rectangle-shaped fields, the mass parameters $m_x$ and $m_y$ control the maximum number of cells the room could occupy in the horizontal and vertical direction.
4) The selection of entry point of each room. The entry point of a room is selected from its boundary cells.

### 2.2.3 Design objectives

The generated space layouts are evaluated by several design objectives which are indicated by a series of fitness values that provide quantitative information about each design solution's performance. The fitness functions for calculating these design objectives are elaborated further below.

1) The internal area of all the rooms in the generated space layout, $A$.
2) The total conflict area (overlapping area) between the generated rooms, $C$.
3) The area of the generated circulation pattern, $L$.

4) The total shadow area in all the habitable rooms (living room, bedroom etc.), $S$.

The total internal area $A$ indicates the usable floor areas in the generated space layout. It is calculated by summing the areas of all the generated rooms in the space layout; $A = \sum_{i=1}^{k} a_i$, where $k$ is the number of rooms and $a_i$ is the area of the $i_{th}$ generated room. As the algorithm optimises the solution candidates to maximise $A$, one naïve strategy would be to simply increase the mass parameters of all the rooms to their upper limits, which could produce floorplans that have large usable floor areas but having all the rooms in irregular shapes. To prevent that, the total conflict area $C$ is introduced as a penalty term. The total conflict area $C$ evaluate the difference between the target area and actual allocated area of each room. The equation for calculating $C$ when applying rectangle-shaped fields is $C = \sum_{i=1}^{k}(m_{xi} * m_{yi} - a_i)^2$, where $m_{xi}$ and $m_{yi}$ are the mass parameters of the $i_{th}$ generated room in the x and y direction. As the target area is calculated from the room's target dimension, which is the maximum area the room could occupy, any overlaps between the rooms would produce a penalty value that grows quadratically. This self-constraint mechanism effectively prevents the optimisation algorithm from increasing the total area $A$ by simply maximising the mass parameters. In order to increase the total area $A$, the optimisation algorithm has to iteratively adjust the positions and mass parameters of the rooms to find proper arrangement of rooms in the design area. These two design objectives are usually applied together during the optimisation as one fitness value $A - C$.

The area of the generated circulation pattern $L$ is calculated by multiplying the total length of the generated pathways and the path width. In this project, the width of the generated pathways is set to 1 meter. The total shadow area $S$ in the habitable rooms is calculated by exposing the generated 3D spatial layout model to direct lighting. This design objective is primarily relevant for apartment space layouts due to their typically fixed envelopes, and for house plans the access to natural light and ventilation is (relatively) easier. Considering that the generative design framework is intended for the schematic design stage and that fenestration often involves aesthetic considerations, we simplify the model by assuming all the external building walls are window walls. The shadow area is calculated through the mesh shadow function of Grasshopper. Figure 10 shows a generated apartment floorplan which has openings in the south and east aspects, the total shadow area in the habitable rooms varies depending on the directions of applied light.

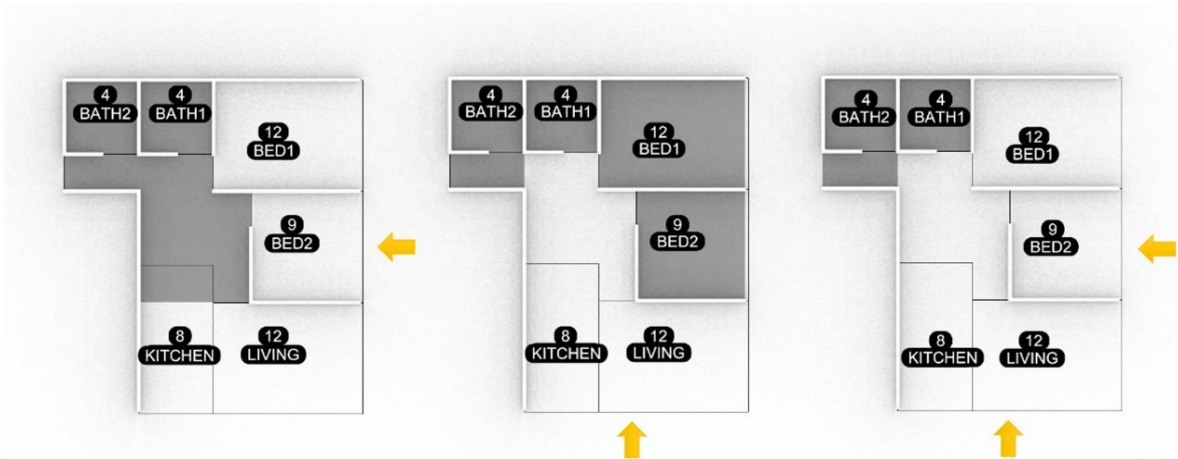

**Figure 10.** A generated apartment space layout and its shadow under different lighting conditions as indicated by the yellow arrows (rendered in Rhino7). The total shadow area in habitable rooms from left to right are respectively 2 $m^2$, 21 $m^2$, and 0 $m^2$.

Figure 11 showcases two apartment space layouts, generated with identical envelopes and room specifications, yet each optimised with building openings in different orientations. In both generated results the habitable rooms are arranged along the window walls to minimise the total shadow area $S$.

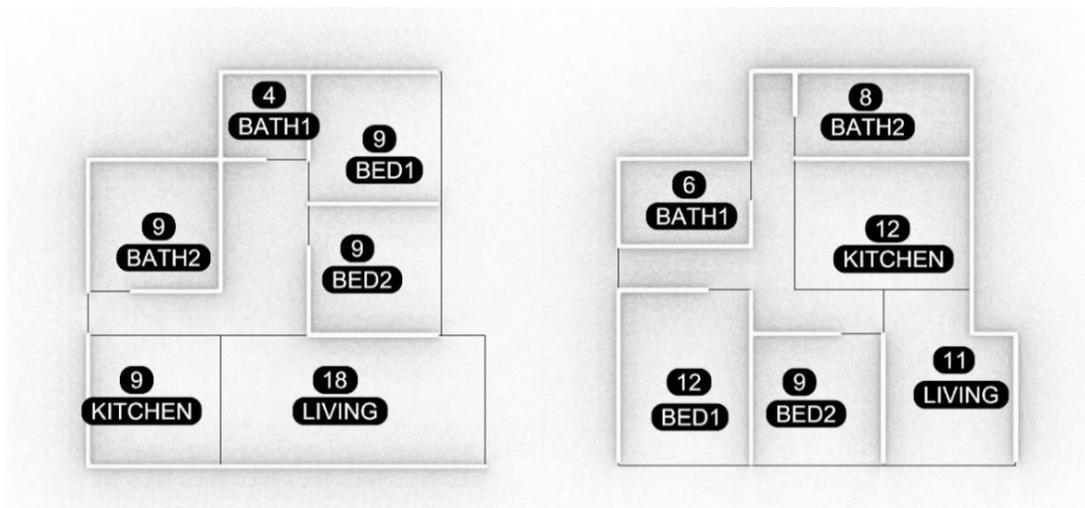

**Figure 11.** Two generated space layouts that optimised with building openings in the east and south aspects (rendered in Rhino7).

## 3. Model application

To assess the effectiveness and practicality of the proposed generative design framework, two distinct case studies were conducted. The first focused on creating floorplans for houses and penthouses—spaces typically more expansive and featuring a greater number of rooms compared to standard apartment units. This case study aimed to evaluate the framework's ability to navigate complex design challenges and its potential to foster creative solutions. The second case study utilised the framework to generate internal space layouts for all the units on the entire floor of a residential apartment building, showcasing the framework's proficiency in handling large-scale design projects. All tests were carried out on a notebook computer equipped with a 2.30 GHz i7-11800H CPU and 16GB of RAM.

### *3.1 The space layout planning of houses*

To demonstrate the generative design framework's capability to generate diverse design solutions, the model was employed to generate space layouts for 4-bedroom houses and penthouses. In comparison with apartment units, houses and penthouses usually have larger design areas and more rooms, which leads to more variations in the generation of space layout and notionally a higher complexity with respect to optimisation. As the design of houses are usually more diverse and creative, the generation of house layouts herein didn't apply fixed building envelopes. Instead, a rectangular envelope working as the block of land was used, which indicates the area and boundary of the generation.

Before generation, the design area inside the envelope was decomposed into grid cells of 1 meter square. Both the house and penthouse were set to include one living area, one dining area, four bedrooms, and two bathrooms. The penthouse featured an additional lift area internally, which indicates the entrance of the whole plan. The generation applied an open-plan design style where the living and dining areas are interconnected and there is no wall separating them.

During the generative design process, three optimisation objectives were employed. The first sought to maximise the overall internal area of the proposed solution, treating room overlaps as a penalty term to encourage the creation of expansive yet regular, rectangular rooms. The second objective aimed to minimise the total area dedicated to circulation, thereby enhancing the connectivity between functional spaces. Additionally, the distance between the living and dining areas was also optimised to maintain a clear functional distinction. As the entire floorplan had access to all external walls of the building, aspects such as lighting and ventilation were not included as optimisation objectives in this generative design process. Notably, increasing the total internal area and reducing the circulation area often present conflicting objectives; maximising internal area alone may result in a design that fills the entire block envelope. Therefore, multi-objective optimisation becomes essential in such generative design endeavours to balance these competing goals effectively.

The Wallacei evolutionary engine was utilised to conduct a three-objective optimisation, configured with a crossover probability of 0.9 and a mutation probability of 0.1. A population size of 50 was established, with the process spanning 100 generations. Out of these, the generative design model performed ten generations, yielding a total of 134 Pareto front solutions. From this pool, six distinctive space layouts were selected for presentation in Figure 12, encompassing four single-story houses and two penthouse designs. The average runtime for the algorithm to complete the generation for each population (50 solutions) was around 40 minutes.

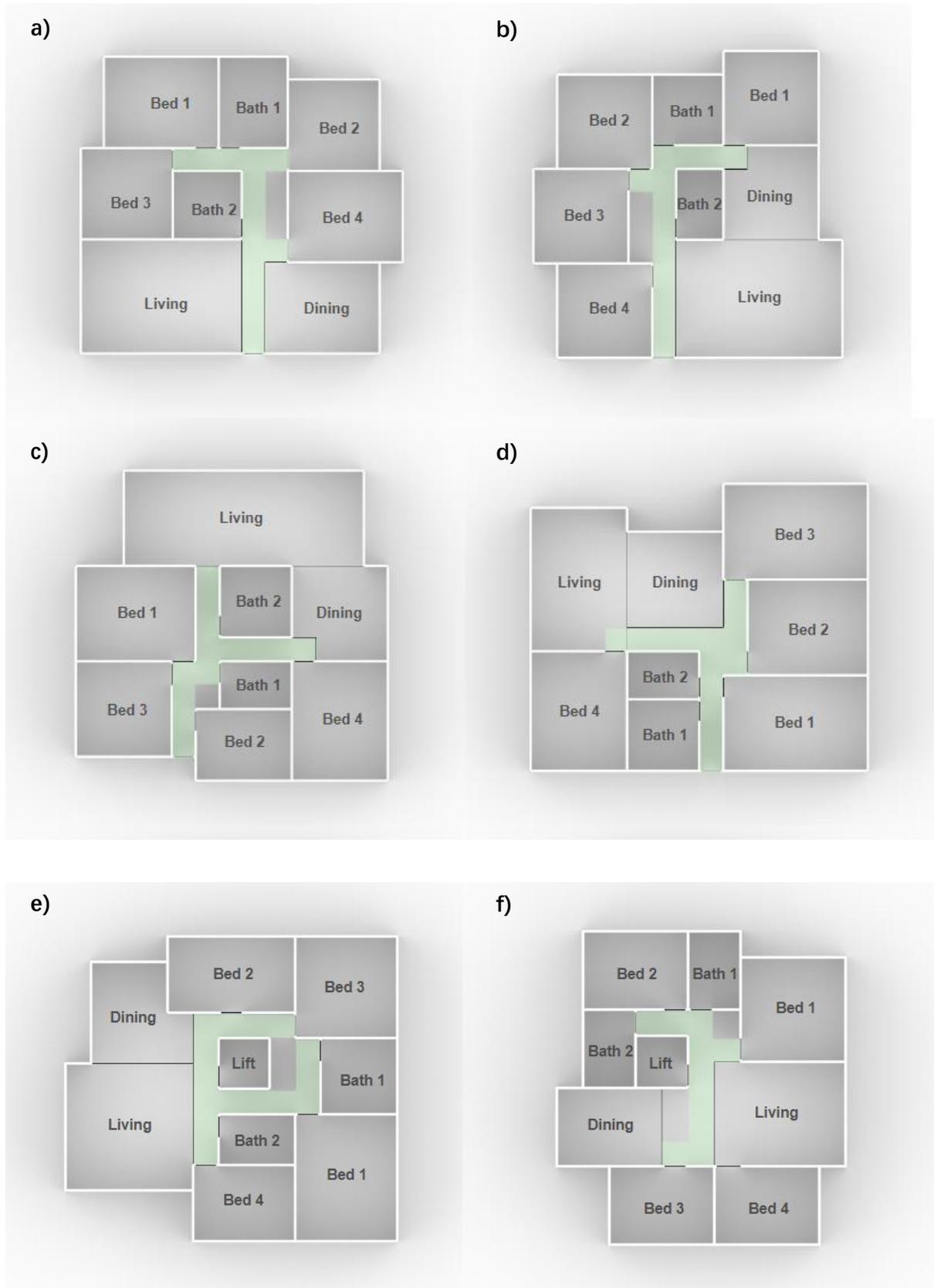

**Figure 12.** Six space layouts of (a-d) a single-story house, and (e-f) a penthouse, automatically generated using the generative design framework developed in this work.

The diversity of the generated floorplans is evident from the figure, where the majority of rooms maintain a regular, rectangular shape. The minor irregularities observed, such as the occasional overlap due to optimisation of the circulation pattern, are minimal. For instance, as seen in Figure 12 (e), the dining area exhibits a conflict area that, due to a stronger field from Bedroom 2, is allocated to that room. The consistent adjacency of all living and dining areas in the presented layouts suggests the effective enforcement of design constraints within the optimisation process. It also showcases the framework's flexibility to accommodate a variety of customised optimisation objectives, catering to specific design needs and preferences. The resulting layouts manifest a cohesive balance of functionality and design efficiency, underscoring the viability of this generative design strategy in creating practical architectural solutions.

### 3.2 The space layout planning of residential apartments

To verify the applicability of the proposed generative design framework in dealing with large-scale design tasks, the proposed model and algorithm were employed to generate space layouts for residential apartment buildings. Specifically, the framework was tasked with creating designs based on the specifications of the 'Evergreen' apartment building project, a residential complex located in Melbourne that features an assortment of apartment styles and configurations [60]. The generation was done with the building envelopes and specifications from floors 5 and 6, which encompass 24 apartment units across 7 different floorplan types. The specific dimensions of these units and floorplan types are detailed in Table 1.

**Table 1.** The dimensions of the apartment units in the floors 5 and 6 of Evergreen apartment in the x and y direction. It is noted that the unit of the dimensions is $u = 2/9$ metre.

| Unit number | Axis | Living | Bed1 | Bed2 | Bed3 | Study | Kitchen | Bath1 | Bath2 | Dining |
|---|---|---|---|---|---|---|---|---|---|---|
| 508-512, 601, 604 | X | 17u | 14u | 14u | | | 16u | 14u | 7u | |
| | Y | 18u | 20u | 17u | | | 14u | 7u | 14u | |
| 503, 505 | X | 18u | 14u | 14u | | | 16u | 14u | 8u | |
| | Y | 16u | 20u | 20u | | | 14u | 9u | 14u | |
| 501, 502, 506, 507 | X | 16u | 14u | | | 14u | 11u | 14u | | |
| | Y | 16u | 17u | | | 8u | 14u | 8u | | |
| 515, 516, 607, 608 | X | 16u | 18u | | | 15u | 14u | 15u | | |
| | Y | 16u | 14u | | | 8u | 12u | 8u | | |
| 517, 518 | X | 16u | 16u | 16u | | | 13u | 12u | | |
| | Y | 11u | 14u | 12u | | | 13u | 7u | | |
| 513, 605, 609 | X | 16u | 16u | 16u | 18u | | 16u | 11u | 14u | 16u |
| | Y | 16u | 24u | 16u | 16u | | 13u | 15u | 9u | 20u |
| 602, 603 | X | 10u | 16u | 16u | | | 11u | 8u | | |
| | Y | 16u | 16u | 16u | | | 13u | 16u | | |

In the original design, the smallest unit of measurement, or 'unit', is defined as $u = 2/9$ metres, and the width of the circulation pattern is $5u$. The floorplan layout design is faithful to dimensions that are a precise multiple of $u$. The original floorplans for levels 5 and 6 of the Evergreen building are illustrated in Figures 13 (top) and 14 (top), providing a reference for the framework's output.

While the minimum unit for generation should (ideally) also be 2/9 metre, practical tests with the generative model revealed that such a fine resolution led to prohibitively high computational costs, due to the generation algorithm's time complexity of $O(N^2)$. Consequently, for this case study, a larger minimum unit of 5u, the same length as the width of the circulation pathway, was adopted for generation. Room dimensions were recalibrated to align with this new unit; for instance, the width ranges for the living room, bedroom (or dining room), study, kitchen, and bathroom were set to [10u, 20u], [15u, 20u], [10u, 15u], [10u, 15u], and [5u, 15u], respectively.

The generation time for a standard two-bedroom unit was approximately 20 minutes, while for the larger units (such as units 513, 605, and 609) with an internal area of 115 $m^2$, the process took about 30 to 40 minutes. In comparison, with the minimum generation unit set to 2/9 meter, the average runtime for each generation was between 2 to 4 hours, due to a significantly expanded search space and reduced likelihood of identifying optimal solutions.

The generation of apartment units also applies an open-plan design style. Given that these generations applied fixed envelopes throughout the process, and considering the significance of natural light and ventilation in the layout of apartment buildings, a two-objective optimisation was implemented. The optimisation goals were to maximise the internal area and minimise the shadow area in habitable rooms, ensuring adequate daylight penetration and ventilation. As mentioned in §2.2.3, the external walls of the apartments were presumed to be entirely made of windows, allowing for the simulation of direct lighting effects on these walls to create mesh shadows in Grasshopper. This simulation aided in the automatic placement of habitable rooms, prioritising light access and ventilation to promote a healthier living environment.

The generative design framework utilised the Wallacei engine to facilitate the generation process, configured with a high crossover probability of 0.9 and a mutation probability of 0.1 to promote diversity. The algorithm was set to run with a population size of 50, iterating through a total of 100 generations. For each unit type within the project, the framework executed 2 generations, culminating in a collection of 167 Pareto optimal solutions for the apartment units. Figures 13 and 14 display a side-by-side comparison of the original and the generated floorplans for levels 5 and 6 of the Evergreen apartment building, providing a visual assessment of the framework's output.

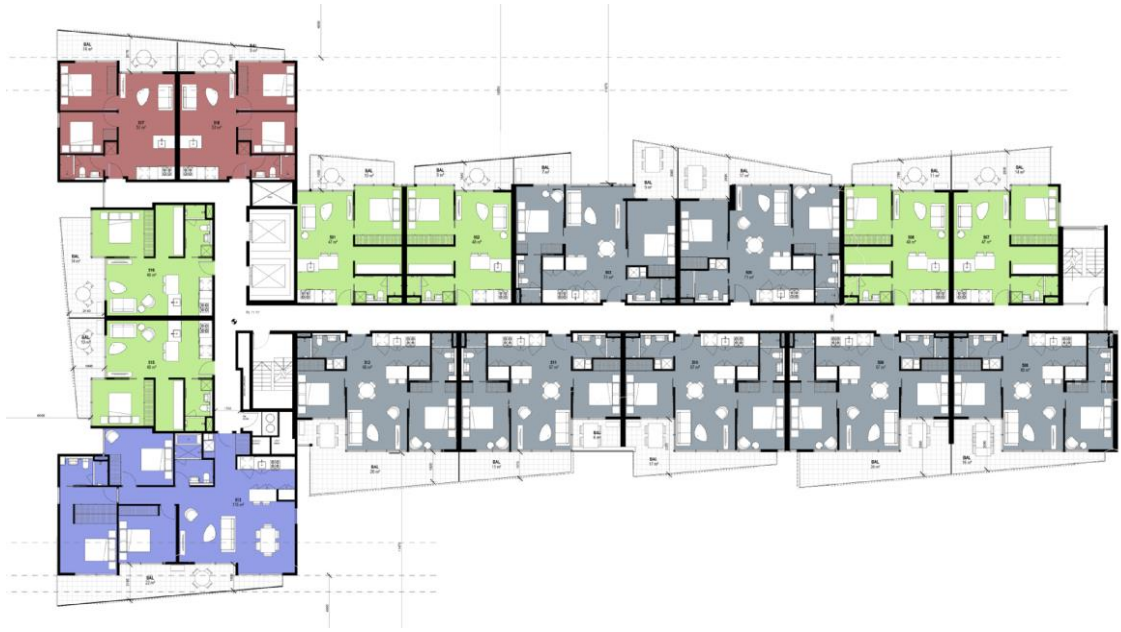
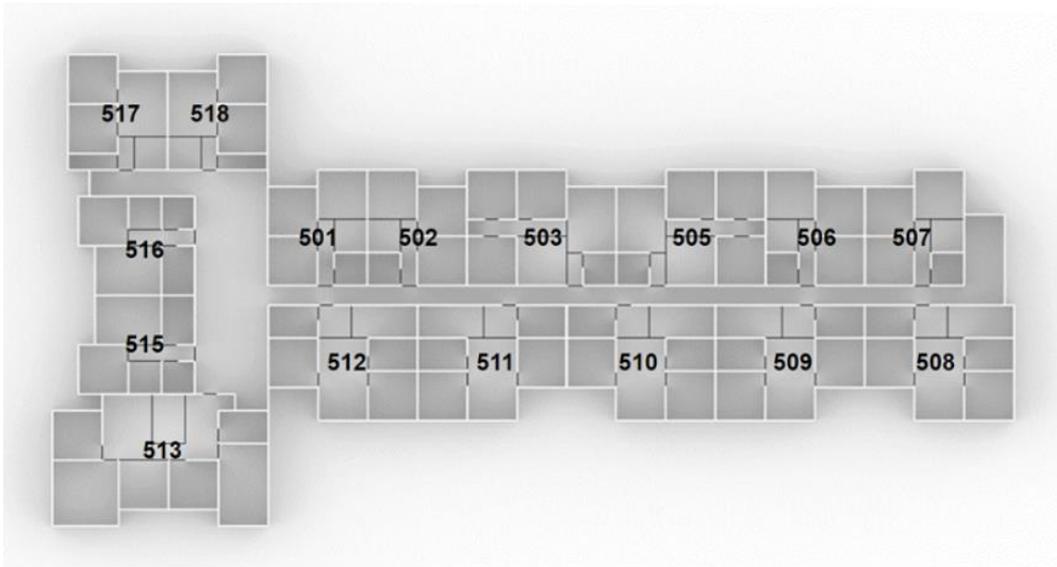
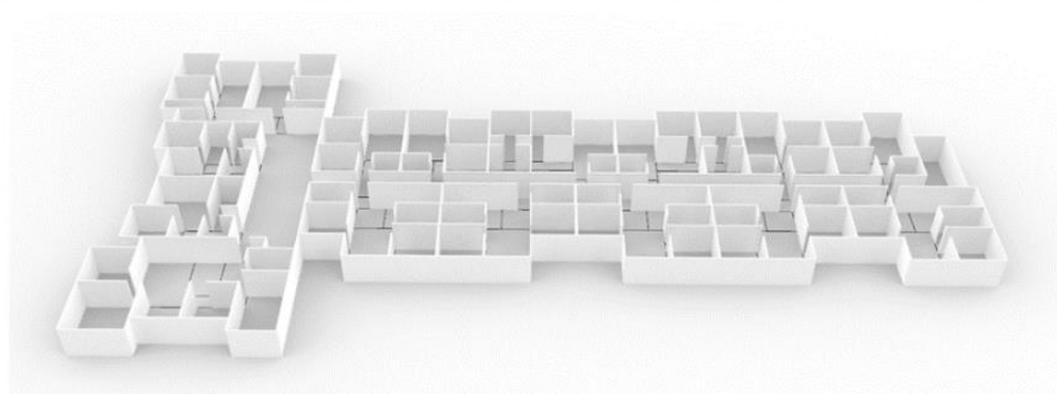

**Figure 13.** The original (top) and generated (middle and bottom) space layout of the level 5 of the Evergreen apartment building (rendered in Rhino 7).

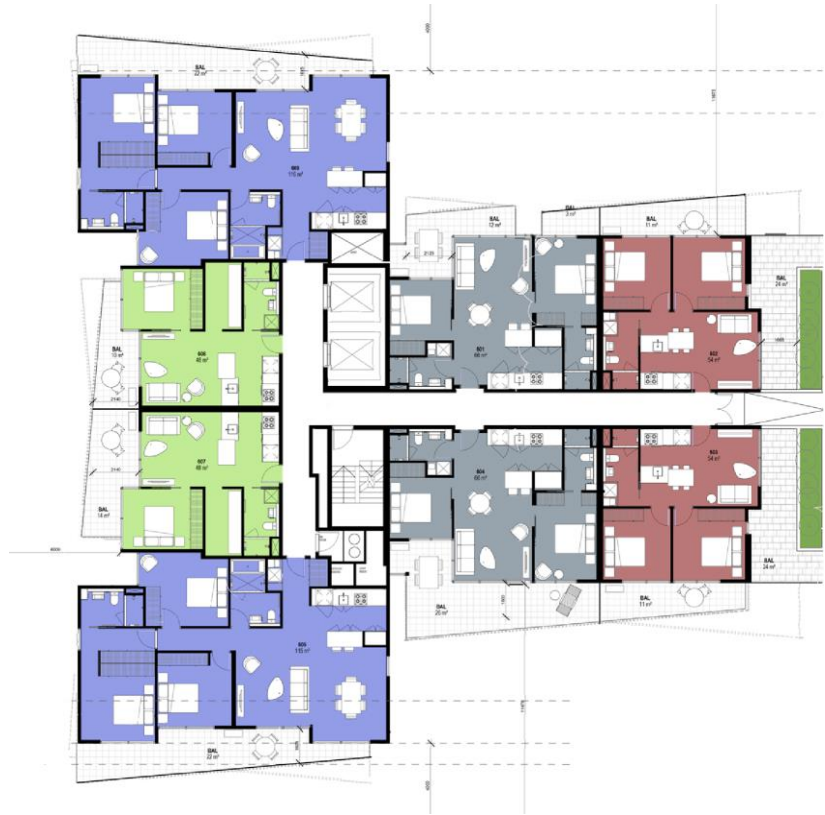
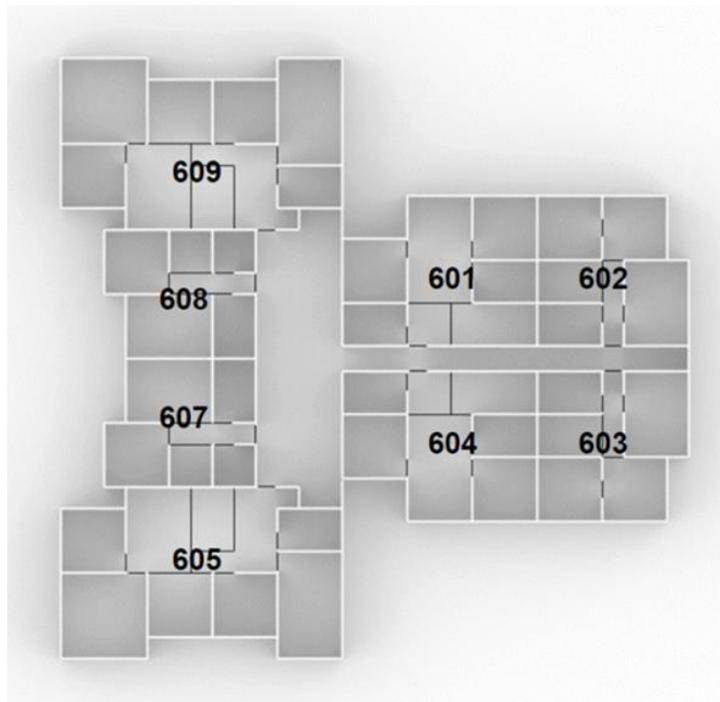

**Figure 14.** The original (top) and generated (bottom) space layout of the level 6 of the Evergreen apartment building (rendered in Rhino 7).

### 3.3 General discussion

The investigation of the generative design framework presented in this paper marks a noteworthy advancement in the realm of architectural design, with a specific focus on the automation of space layout generation. This framework, through its robust capacity to create a variety of optimised floorplans, has demonstrated potential as a tool for both conceptual exploration and practical application. Central to our model is an innovative physics-inspired parametric approach, and unlike many methods discussed in the literature, it incorporates a mechanism for generating circulation patterns, a critical component in ensuring effective design solutions.

When juxtaposed with conventional methodologies such as cellular automata, our model exhibits a greater capacity to address complex design challenges. Cellular automata are typically limited by simplistic local rules that are not well-suited to complex forms. In contrast, the proposed parametric model in our study can manage intricate design requirements and offer nuanced control over the outcomes. Compared to shape grammar, which is restricted by rigid generative rules that can inhibit design variation, our parametric model is more flexible and customisable, readily tailoring to a wide range of design scenarios and specific client needs.

A significant advantage of our parametric model is its compatibility with advanced evolutionary algorithms, such as NSGA-II. This facilitates an effective and sophisticated search for optimal designs within a multi-objective landscape. The potential for further integration with other optimisation algorithms—such as particle swarm optimisation (PSO), simulated annealing (SA), and ant colony optimisation (ACO)—opens avenues for achieving diverse optimisation and design goals.

Developed within a professional design software environment, our model's optimised parametric outputs can be directly linked to visualisation tools. This allows for immediate feedback on how parameter adjustments impact the overall design, providing a dynamic and interactive design experience. The scalability of our parametric models is also notable; they are applicable to both small-scale initiatives, like individual apartment units, and extensive development projects, demonstrating their broad utility.

When comparing the parametric model used in our framework with machine learning methods such as Generative Adversarial Networks (GANs) and Variational Autoencoders (VAEs) for creating floorplans, several distinct advantages of our approach become apparent. Although generative ML models can quickly provide design solutions once trained, they rely on extensive datasets and generally present their outcomes as pixel representations within images.

This reliance on data introduces several challenges in ML-based generative models. Firstly, the outcomes are highly dependent on the dataset used, and the generation process is largely random, leading to difficulties in customising results. Moreover, the database might exhibit a bias toward certain design styles. Secondly, the comprehensiveness of the dataset influences the feasibility of generated solutions, with a limited dataset potentially resulting in impractical designs. Additionally, image-based outputs from these models often display cluttered and unclear boundaries between rooms. This lack of clarity makes it challenging to translate ML-generated results into parametric models used in professional architectural software without undergoing a complex and time-consuming conversion process.

In contrast, the parametric model proposed in this study operates on a set of explicit rules and parameters to generate designs, enabling the creation of diverse and innovative design layouts that might not be represented in any existing dataset. The outcomes are inherently parametric, ready for immediate application in professional design environments, offering the ability to be tailored and refined to meet the specific demands and constraints of individual projects.

However, it's crucial to acknowledge the limitations inherent in this approach. The generative framework, while powerful, still requires fine-tuning to fully encapsulate the nuances of architectural aesthetics and functionality that a human designer naturally considers. For instance, the algorithm may produce designs that deviate from established architectural norms, indicating areas where further development is necessary. While these solutions may initially appear unconventional or 'not right', they can serve as creative prompts that challenge traditional design thinking and potentially reveal novel solutions that might be missed by a human designer.

Another limitation is the computational complexity associated with generating floorplans. Given that the computational complexity of each design solution is $O(N^2)$, employing higher resolutions leads to a quadratic increase in computational demand. This relationship between resolution and computational intensity can impose practical constraints on the model's application, particularly when dealing with large-scale projects or when striving for high-resolution details in the design outputs. Addressing this computational challenge is essential for scaling the use of the generative design framework to a wider array of architectural projects while maintaining a balance between detail and efficiency.

Looking ahead, there is substantial scope for extending the capabilities of this framework. Future work could integrate more sophisticated aesthetic evaluation criteria, enabling the algorithm to better judge the qualitative aspects of design. Incorporating considerations for environmental sustainability metrics, such as energy efficiency and material conservation, could also enhance the practical utility of the generative design framework. There is also potential to refine the algorithm to better address the specific needs of different architectural styles and cultural contexts, broadening its applicability.

Another exciting avenue for advancement involves the synergy of machine learning methods with this framework. The current model operates independently of any existing datasets and domain knowledge, relying solely on predefined objectives and parameter ranges to generate and optimise design solutions within a given framework. The optimisation algorithm embarks on its search from a randomly generated population of design solutions. However, by introducing machine learning methods that learn from a vast array of existing designs, the generative design process might be significantly enhanced.

The trained machine learning model can be utilised to generate preliminary design proposals, providing a more informed and data-driven starting point for optimisation. Transitioning from an initial population that is entirely random to one augmented by machine-learned proposals holds the potential to enhance the efficiency of identifying optimal solutions. Such a hybrid approach could streamline the entire optimisation workflow, yielding high-quality, varied, and data-informed design solutions that merge computational creativity with learned architectural patterns. This synergy could be particularly beneficial in navigating the complex landscape of modern architectural design, where both innovation and adherence to established practices are valued.

Addressing the challenge of computational complexity, another direction for future work could involve segmenting the optimisation process. Given that the efficiency of the optimisation is influenced by the resolution used, splitting the generative design process into phases could be a viable solution. This approach would initially conduct optimisation at a lower resolution to determine the approximate positions and dimensions of each room. This stage focuses on the macro layout, streamlining the decision-making process by reducing the computational burden associated with high-resolution optimisation. Once this broader layout is established, the results can then be scaled up to a higher resolution for fine-tuning. At this stage, detailed adjustments to individual rooms and their relationships with surrounding functional spaces can be made. This two-phase approach allows for more nuanced and detailed design refinement while minimising the generation of ineffective design solutions in the early stages, thus conservatively using computational resources. By adopting this segmented strategy, the framework can efficiently navigate the initial broad layout decisions before delving into the intricacies of finer spatial arrangements. This method promises to balance the need for detailed, high-quality design outcomes with the practical constraints of computational resources and time efficiency.

# 4. Conclusions

A generative design framework was presented, that combines a novel physics-inspired parametric design model and an evolutionary optimisation method to generate the internal space layouts of buildings. For a set of given design constraints and objectives, the proposed generative model could automatically generate internal space layouts. By applying a field-mimicking space allocation mechanism and an automatic circulation generation algorithm, the proposed parametric model could be applied to search and explore a wide variety of feasible internal layout designs - for building envelopes of various forms. The proposed generative design method does not follow a traditional design workflow and design conventions, instead, it maintains a collection of design solutions and finds optimal designs through a stochastic multidirectional optimisation process. A multi-objective evolutionary optimisation method was applied to evaluate and optimise the design solutions against a set of quantitative design objectives.

Two study cases were presented to examine the applicability of the proposed generative design model. In the first case study, the model was employed to generate internal layouts for single-story houses and penthouses, applying non-fixed building envelopes. The aim was to test the model's capability in dealing with complex design problems. To verify the applicability of the proposed generative design framework in dealing with large-scale design tasks, in the second case study, the model was applied to generate internal layouts for the space layouts of two floors of a residential apartment building with 24 apartments of various size and configuration. From both study cases, it shows that the proposed generative design model could be applied to generate a variety of feasible design solutions for design problems of various design constraints.

Nonetheless, there several limitations that persist within the current framework. The current approach can be computationally intensive, particularly when high-resolution design spaces are required. Future work could focus on improving computational efficiency, perhaps by segmenting the optimisation process or by integrating machine learning methods to infuse domain expertise into the generative design workflow. Additionally, further development could aim to enhance the model's sensitivity to aesthetic criteria and integrate environmental sustainability considerations, broadening the scope and applicability of the framework.

In conclusion, the generative design framework presented in this paper demonstrates significant potential as a transformative tool in the field of architectural design. Its capacity to generate diverse and optimised solutions heralds a paradigm shift in the automated design of building interiors, setting a precedent for more innovative, efficient, and responsive approaches to architectural practice. Although this model was originally conceived for the design of architectural interiors, the generative design workflow it embodies has the versatility to be applied to other domains involves spatial layout design and optimisation.

**Acknowledgements**

The authors gratefully acknowledge Hayball Architecture and Robert Marcen for technical discussions.